%% file: camera_ready.tex
\documentclass[10pt,twocolumn,letterpaper]{article}

\usepackage{cvpr}              

\usepackage{times}  
\usepackage{helvet}  
\usepackage{courier}  
\usepackage[hyphens]{url}  
\usepackage{graphicx} 
\usepackage{natbib}  
\usepackage{caption} 
\usepackage{algorithm}
\usepackage{algorithmic}
\usepackage{multirow}
\usepackage{booktabs}
\usepackage{amsmath}
\usepackage{graphicx}
\usepackage{subcaption} 
\usepackage{newfloat}
\usepackage{listings}

\input{preamble}
\definecolor{cvprblue}{rgb}{0.21,0.49,0.74}
\usepackage[pagebackref,breaklinks,colorlinks,allcolors=cvprblue]{hyperref}

\def\paperID{15523} 
\def\confName{CVPR}
\def\confYear{2026}

\title{Towards Lossless Ultimate Vision Token Compression for VLMs}

\author{Dehua Zheng, Mouxiao Huang, Borui Jiang, Hailin Hu, Xinghao Chen\\
Huawei Noah’s Ark Lab\\
{\tt\small \{zhengdehua2, huangmouxiao, jiangborui, hailin.hu, xinghao.chen\}@huawei.com}
}

\begin{document}
\maketitle
\input{sec/0_abstract}    
\section{Introduction}

In recent years, the extensive deployment of large language models (LLM) in visual domains has catalyzed significant progress in visual language models~(VLM)~\cite{llava_next,qwen25_vl,internvl2_5,gpt_4}. VLM typically adopts a framework comprising visual encoder~\cite{CLIP,siglip} and projector to achieve cross-modal alignment between textual and visual domains. Consequently, VLMs have demonstrated robust performance across diverse downstream applications, encompassing visual question answering, spatial localization, and temporal reasoning tasks. However, VLMs suffer from prohibitive inference costs due to their massive parameterization (often in the billions) and quadratic computational complexity with respect to input sequence length. The sequence length of high-resolution images and videos further exacerbates this issue~\cite{FastV}, and becomes the primary bottleneck. These constraints severely limit the real-time deployment of VLMs in practical applications.

\begin{figure}[htbp]
	\centering
	\begin{subfigure}{0.22\textwidth}
		\includegraphics[width=\linewidth]{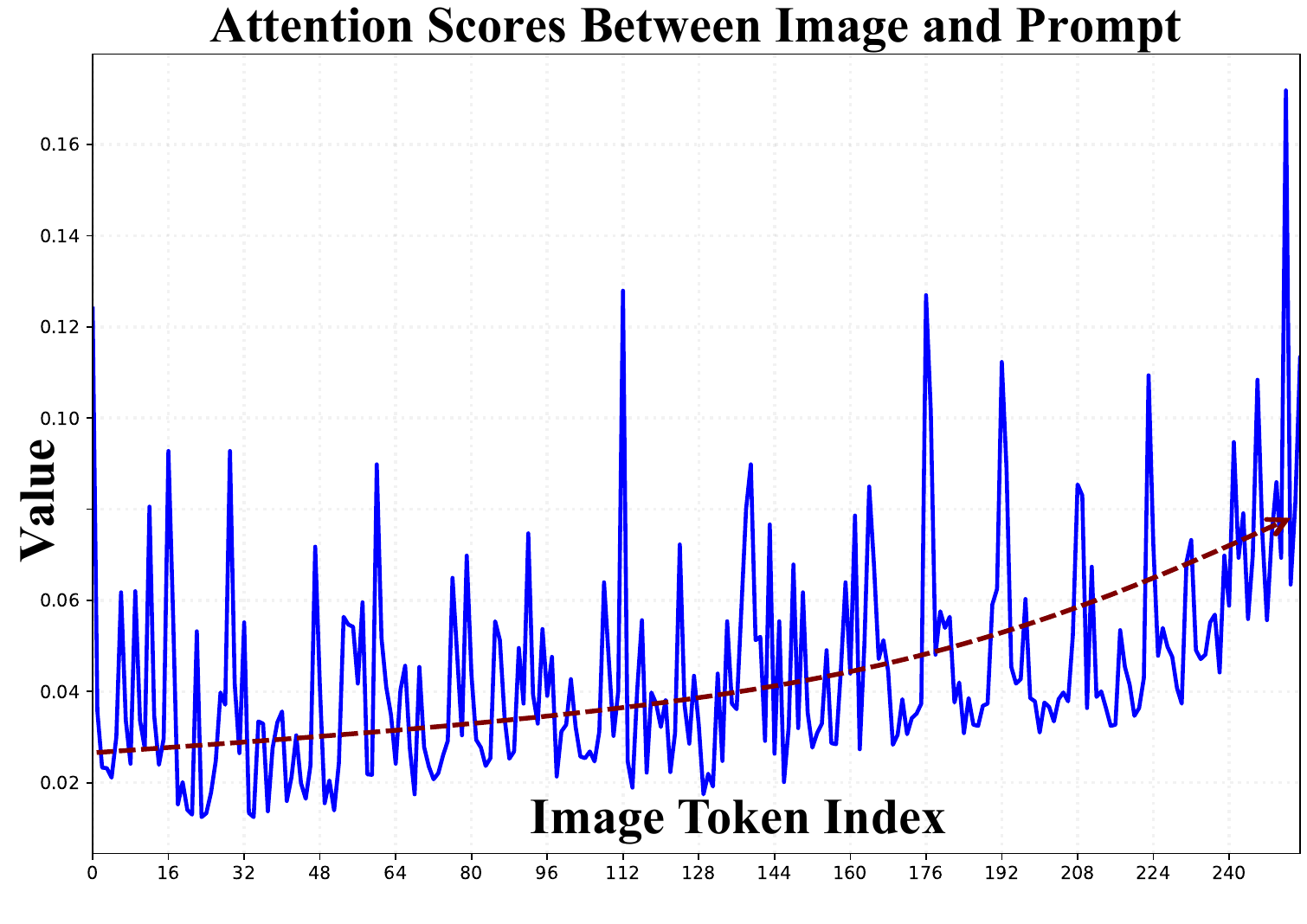}
		\caption{Image-text score.}
		\label{fig:layer1}
	\end{subfigure}
	\hfill 
	\begin{subfigure}{0.22\textwidth}
		\includegraphics[width=\linewidth]{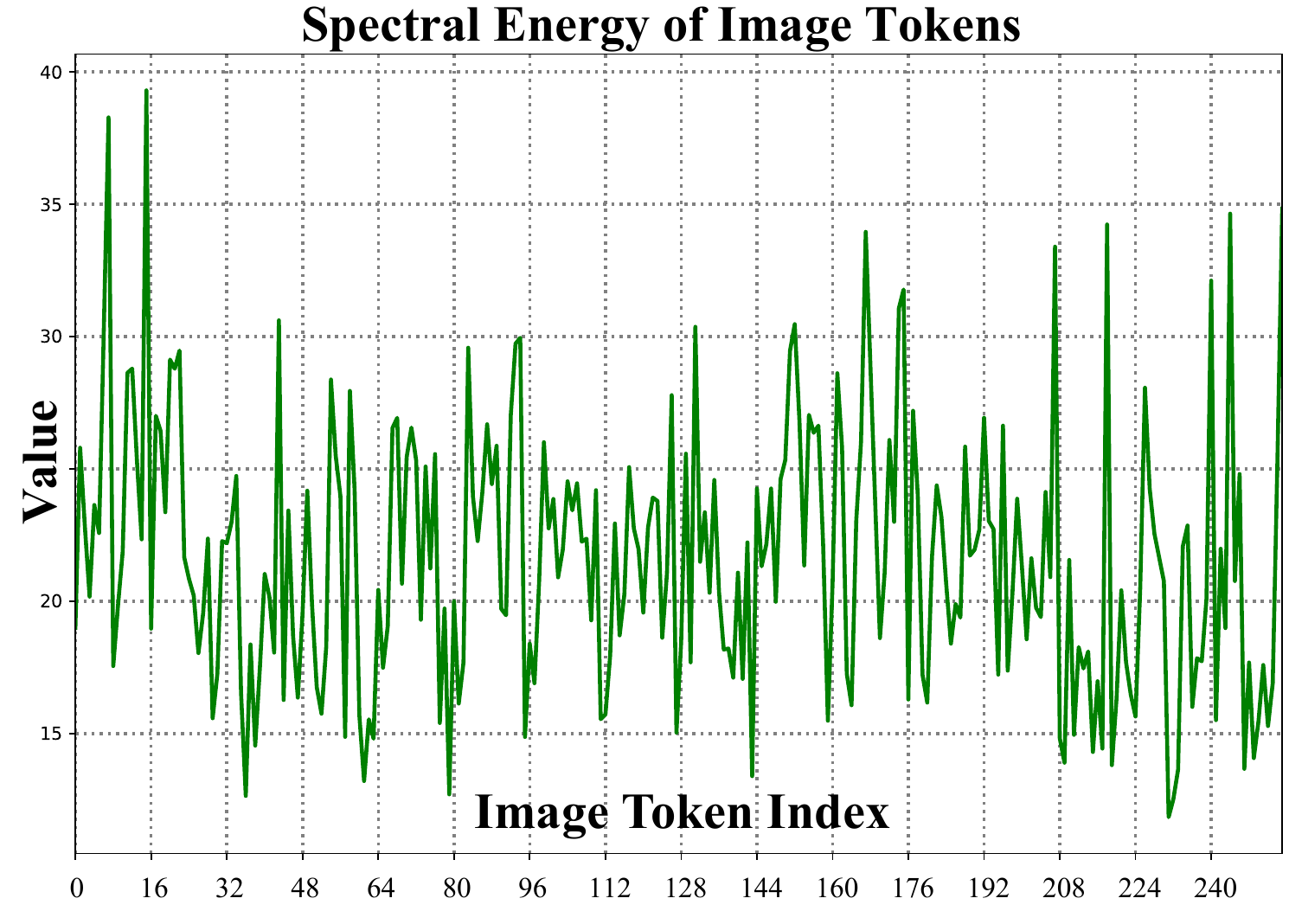}
		\caption{Image tokens spectrum.}
		\label{fig:layer2}
	\end{subfigure}
	\caption{The distribution of image-text attention scores and tokens spectrum. Due to position bias, the attention scores exhibit a significant upward trend near the text region, whereas the spectrum curve remains position-invariant.
	}
	\label{fig:token_distribution}
\end{figure}

Extensive research has been devoted to compressing visual tokens to mitigate redundancy and accelerate the inference process of LLM. These approaches fundamentally operate either at projector or within the LLM, eliminating or merging redundant visual tokens based on similarity metrics or attention score ranking mechanisms.  We systematically examine these approaches through two methodological foundations while critically evaluating their inherent limitations. (1) Attention-aware pruning. These approaches primarily employ either special token~(eg. [CLS]) or text-guided attention to rank the importance of visual tokens. For example, FastV~\cite{FastV} and SparseVLM~\cite{SparseVLM} develop text-guided compression algorithms based on cross-modal attention between visual and textual tokens. However, these methods suffer from significant position bias~\cite{fastervlm,we_right}. As illustrated in Fig.~\ref{fig:layer1}, position bias introduces substantial noise in importance estimation, tokens adjacent to text consistently exhibit inflated attention weights regardless of their semantic relevance. Furthermore, due to reliance on explicit attention score, these compression methods cannot be accelerated using modern FlashAttention~\cite{flash_attn} optimization techniques. (2) Similarity-aware merging.  These methods typically compute the similarity between tokens and employ bipartite matching~\cite{ToMe,visionzip,TinyChart} or clustering-based~\cite{chat_univ,pact} approaches for vision token compression. However, their computational efficiency significantly deteriorates when processing longer vision token sequences. Additionally, these compression algorithms focus solely on local token-wise proximity while neglecting the global information distribution. They tend to excessively merge tokens in dense regions while inadequately merging them in sparse regions, thereby compromising semantic integrity and class balance, leading to distributional shift~\cite{ToFu}. Moreover, clustering-based approaches disrupt the original positional arrangement of tokens, resulting in disordered spatial information. 

Additionally, early-layer LLM pruning risks semantic degradation, forcing most methods to only prune deeper LLM layers while keeping visual encoder, projectors and early LLMs intact. Despite attempts~\cite{multi_stage} to integrate a limited number of visual tokens in the visual encoder, existing methods compromise spatial structur and remain incompatible with versatile 2D projector architectures. Thus, token compression across the entire VLM remains largely unexplored. 

\begin{figure}
	\begin{center}
		\includegraphics[width=8.2cm]{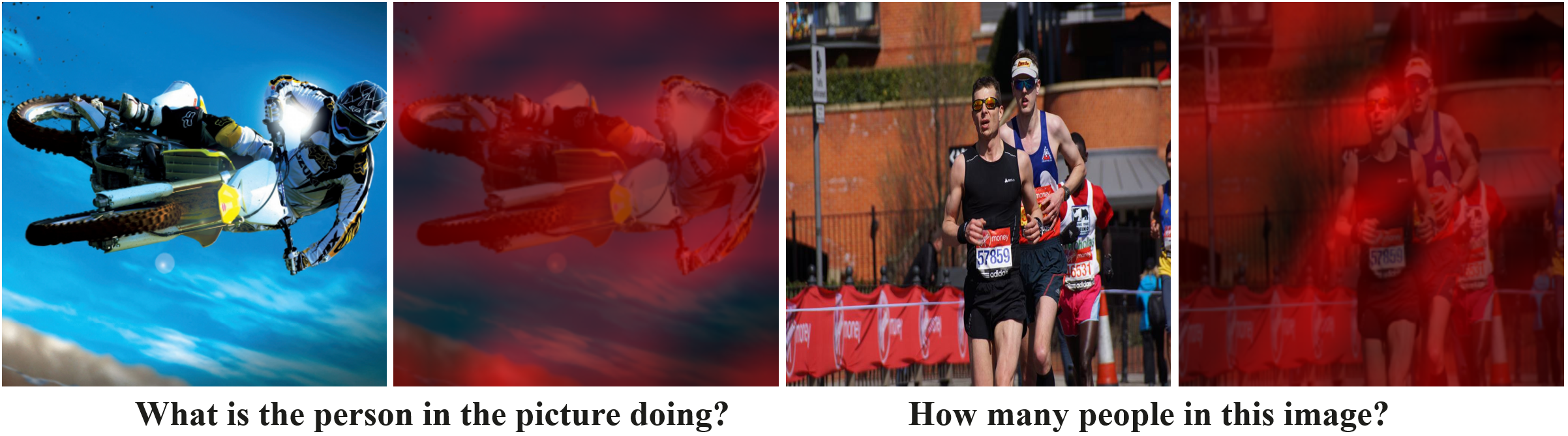}
	\end{center}
	\caption{Visualization of low-frequency visual tokens of Internvl2.5-8B in layer-8.}
	\label{fig:spectrum_visual}
\end{figure}

To overcome the above limitations, we propose a more robust and resilient algorithm towards \textbf{L}ossless \textbf{U}ltimate \textbf{V}ision token \textbf{C}ompression~(\textbf{LUVC}), which operates across the entire network and is training-free. LUVC extends token compression to the visual encoder through a computationally efficient \textbf{O}rthogonal \textbf{I}terative \textbf{M}erger~(\textbf{OIM}) and eliminates noisy tokens through a \textbf{S}pectrum \textbf{P}runing \textbf{U}nit~(\textbf{SPU}) equipped with a carefully crafted low-pass filter in LLM. As shown in Fig.~\ref{fig:layer2}, visual token spectrum exhibits robustness against position bias, which provides reliable grounding for stable pruning unlike attention weights. Existing researchs~\cite{Fourier_Domain,Spectrum_Preserving} have empirically verified the inherent high-frequency attenuation property in Transformer architectures during forward propagation, providing additional theoretical support for our frequency-based SPU. As shown in Fig.~\ref{fig:spectrum_visual}, visualization reveals that low-frequency features exhibit a concentrated distribution in visually salient regions, which further corroborates the feasibility of frequency-based pruning. For the compression during visual encoder, OIM performs iterative merging that is orthogonal in spatial. OIM maintains the intrinsic spatial structure of visual inputs, ensuring compatibility with various projector architectures, while simultaneously increasing parallelizability for more computationally efficient processing. Our contributions are as follows: 

\begin{itemize}
	\item Based on the spectrum characteristics of Transformer features, we designe the attention/similarity-free spectrum pruning unit~(SPU), which gradually prune high-frequency noise tokens until all visual information is fused into the multimodal queries, achieving ultimate pruning of visual tokens at the final LLM layer.
	
	\item We design the orthogonal iterative merger~(OIM) that merges tokens in spatial axes to reduce the computation across the entire VLM. 
	OIM maintains the saptial structure of visual input and is suitable for diverse projectors.

	\item We integrate the above two strategies to construct training-free LUVC, achieving approximately 2× acceleration while maintaining nearly lossless performance. Experimental results confirm that the plug-and-play LUVC demonstrates remarkable compatibility across diverse VLM architectures and parameter scales, while achieving superior performance on diverse tasks including single image, multi images, videos, and particularly information-dense chart and document processing.
	
\end{itemize}

\section{Related Work}

\subsection{Multimodal Large Language Models}
VLM architectures have diversified in design. Taking projectors as an example, the BLIP~\cite{blip,blip_2} and  MiniCPM-V2.6~\cite{miniCPM} employ cross-attention mechanisms. The LLaVA series~\cite{llava,llava_next} introduce a two-layer MLP to bridge visual and language models. InternVL2.5~\cite{internvl2_5} and Qwen2.5-VL~\cite{qwen25_vl} employ pixel shuffle and token merger modules, respectively, which necessitates preservation of spatial structural information. The evolutionary trajectory of these models has progressed from rudimentary image comprehension to encompass a broad spectrum of downstream applications. This expansion has engendered specialized model variants, including video-oriented architectures~\cite{video_llava,intern_video,Video_Poet} and document analysis systems~\cite{doc_owl,tiny_chart}, each tailored to their respective domains. Gemin~\cite{Gemini} and LWM~\cite{WMM} have established long-context modeling as a fundamental capability. This theoretical advancement has catalyzed significant progress in high-resolution image and long-form video understanding. This paradigm presents critical challenges for model efficiency, mandating the development of sophisticated token compression methodologies.

\begin{figure*}
	\begin{center}
		\includegraphics[width=15.6cm]{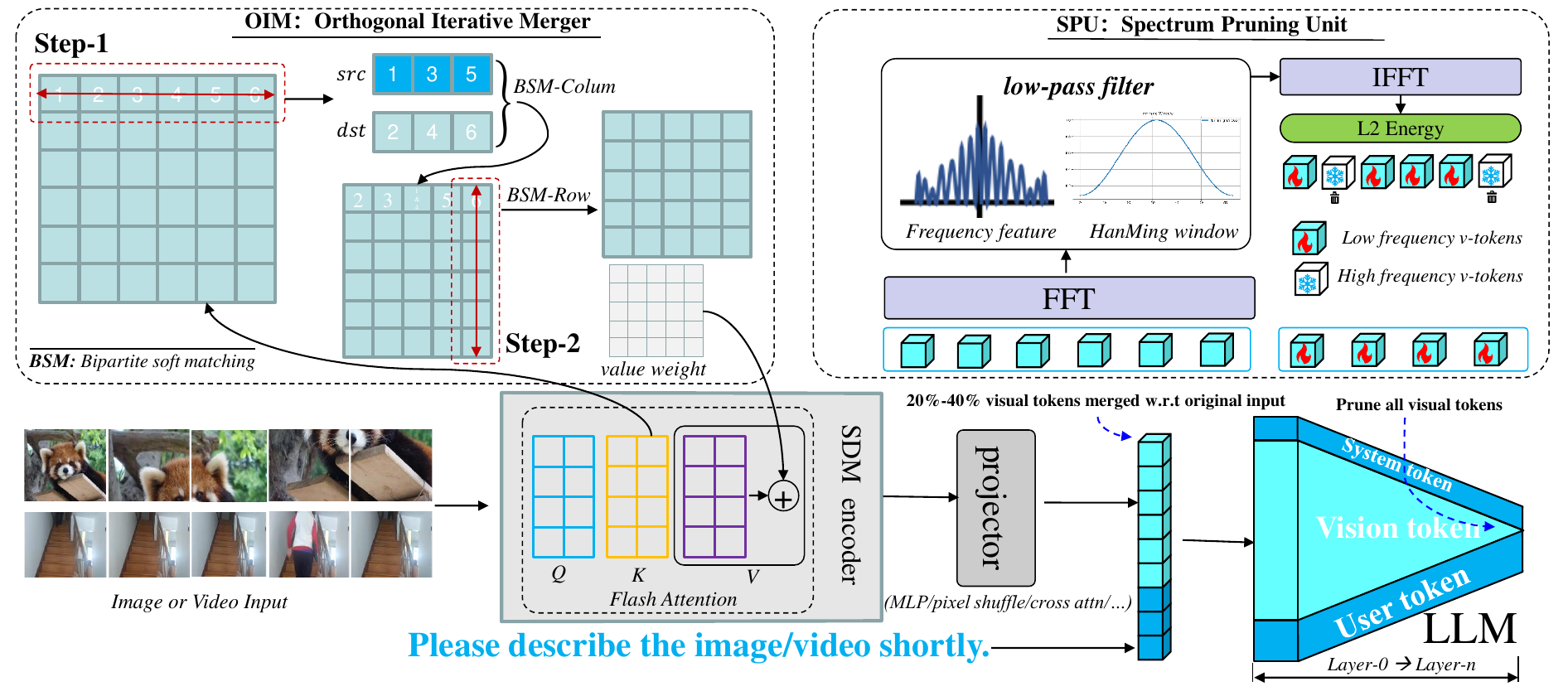}
	\end{center}
	\caption{The overview of LUVC. LUVC consists of two key components. (1) Orthogonal Iterative Merger~(\textbf{OIM}), which maintains the spatial structure while increasing computational parallelism by performing step-by-step merging in both lateral and longitudinal dimensions. (2) Spectrum Pruning Unit~(\textbf{SPU}), which applies low-pass filtering to visual tokens through FFT and IFFT,  which achieves progressive token pruning via cascade structure.
	}
	\label{fig:overview}
\end{figure*}

\subsection{Vision Token Compression Methods}
Numerous token pruning techniques, inspired by traditional visual transformers pruning~\cite{evo_vit,dynamic_vit,focus_detr}, or novel token compression schemes for VLMs, have been proposed to accelerate inference. FastV~\cite{FastV} was the first to explore redundant visual tokens using text-guided attention score. FasterVLM~\cite{fastervlm} introduced an alternative approach by utilizing the attention of the [CLS] token. SparseVLM~\cite{SparseVLM} further improved upon FastV by performing an initial screening of visual queries and recovering pruned visual tokens. Some works have adopted various advanced techniques during the training phase. For instance, Qwen2-VL~\cite{qwen2_vl} adopts a token merging strategy during training, where four spatially adjacent patches are aggregated into a single consolidated visual token. MiniCPM-V2.6~\cite{miniCPM} introduces trainable queries, which dynamically projects  visual segments into fixed-dimensional latent representations. While these approaches require sophisticated end-to-end training frameworks with non-trivial computational overhead, and limit applicability exclusively to proprietary model, precluding generalization to open-source or third-party VLMs. FastVLM~\cite{FastVLM} innovatively integrates FastViTHD to maintain high-resolution image processing capabilities while producing fewer output tokens. VTW~\cite{VTW} demonstrated that visual tokens can be aggressively pruned at the final stage of LLMs, a finding validated on several simple vision benchmarks. Additionally, various clustering-based methods merge similar tokens based on semantic similarity and static distance, such as Chat-UniVi~\cite{chat_univ}, PACT~\cite{pact}, and LLaVA-PruMerge~\cite{LLaVA_PruMerge}. However, due to inherent limitations, these algorithms cannot achieve optimal performance. To overcome these limitations, we designed an innovative pruning framework LUVC through a spectrum pruner and a orthogonal iterative merge.

\section{Methodology}
In this section, we provide a detailed introduction to the LUVC method, as illustrated in the Fig.~\ref{fig:overview}. This method achieves the complete elimination of visual tokens by merging similar tokens during the visual encoder and gradually filtering out high-frequency tokens during the LLM, ultimately integrating all low-frequency, high-information visual tokens into the multimodal query. The LUVC consists of two core components. First, a orthogonal iterative merger is designed during the visual encoder, performing iterative merging in spatial axes. Then, in the LLM phase, the merged visual tokens undergo spectrum analysis through Fast Fourier Transforms~(FFT) and Inverse Fast Fourier transforms~(IFFT), with a carefully designed low-pass filter progressively filtering out high-frequency noise tokens. Notably, spectrum pruning begins at an earlier layer of the LLM and is cascaded to cover the final layer. Section 3.1~\ref{sec:3_1} provides a priori analysis, discussing the computational complexity of VLMs, and the importance of low-frequency visual tokens. Section 3.2~\ref{sec:3_2} analyzes the design of the orthogonal iterative merger, while Section 3.3~\ref{sec:3_3} details the design of the spectrum pruning unit and low-pass filter.
\subsection{Preliminary Analyze}
\label{sec:3_1}
\textbf{Computational Complexity Analysis.} The $Flops$ of VLMs primarily arise from the self-attention~(SA) and feed-forward network~(FFN). We define the $Flops=O(n^2d+nd^2)$, where $n=v_t+n_v$ is the total token sequence length and $d$ is the hidden dimension. For InternVL-2.5, a frame or image patch generates $n_v=256$ visual tokens. For a video tith 32 frames, $n_v\approx 8\times10^3$, which is approximately two orders of magnitude greater than $n_t$. This excessive redundancy of $n_v$ imposes substantial computational overhead during visual encoder and LLM. Meanwhile, we analyze the $Flops'=O(n_v^2k)$ of traditional merge strategies, where $k$ is the dimension of keys. In contrast, our OIM perform orthogonal merging in spatial axes. The visual tokens is transfered from ${R}^{1\times n_v\times k}$ to ${R}^{\sqrt{n_v}\times \sqrt{n_v}\times k}$, $Flops'=O(n_v^{1.5}k)$, which is computationally more efficient and maintains the saptial structure.

\textbf{The Importance of Low-frequency Visual Tokens.} Due to the global attention mechanism of Transformers, low-frequency visual tokens more extensive global and cross-modal information, which is more crucial for high-level semantics. In addition, low-frequency visual tokens are more robust to noise and minor disturbances, which avoids pruning errors caused by noise (such as translation, lighting changes).  Numerous works~\cite{Fourier_Domain, Fourier_Perspective} have rigorously derived the mathematical characterization of this preference for low-frequency signals in Transformer. We define visual token $z\in\mathbf{R}^{D}$ and fourier feature $\sum_{n=0}^{N-1}ze^{-i2\pi kn/N}$, $DC[z](k=0)$ is Direct-current Component~(DC) and $H[z](k=1,2,...N-1)$ is High-frequence Component~(HC), $A$ is the attention-weights. Formula~\ref{equa:equa_1} is a straightforward result of Perron-Frobenius theorem, which demonstrates that Transformer exhibit an inherent bias toward low-frequency components and treat high-frequency signals as noise or redundancy to gradually smooth them out.  

\begin{equation}
	\label{equa:equa_1}
	\lim_{t \to \infty}\frac{||HC[A^tz]||_2}{||DC[A^tz]||_2}=0 ~\cite{Fourier_Domain}
\end{equation}


In summary, low-frequency visual tokens contain more critical semantics and are less susceptible to noise interference, a theory further substantiated by Fig.~\ref{fig:spectrum_visual}.

\begin{figure}
	\begin{center}
		\includegraphics[width=8.2cm]{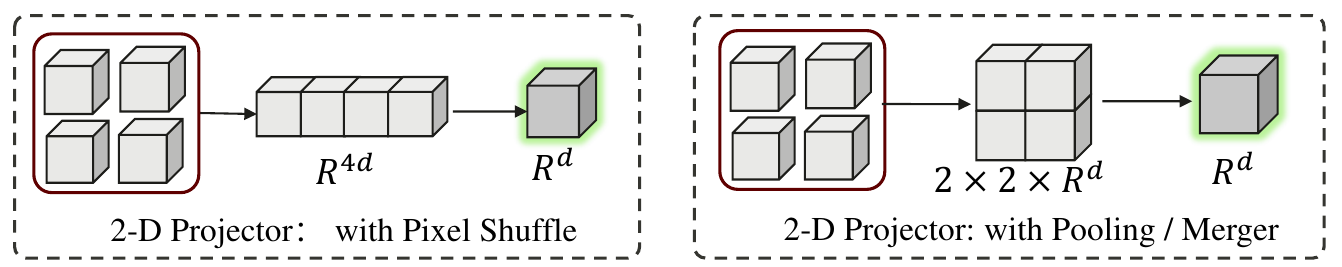}
	\end{center}
	\caption{The examples of 2D projectors demonstrate that they rely on the 2D spatial structure.
	}
	\label{fig:1D_ToMe}
	\vspace{-1.8em}
\end{figure}

\subsection{Orthogonal Iterative Merger}
\label{sec:3_2}
Merging in one-dimensional space disrupts the spatial structure of the original visual input. As shown in the Fig.~\ref{fig:1D_ToMe}, many contemporary projector architectures operate natively in 2D spatial structure~\cite{qwen25_vl,internvl2_5}, fundamentally precluding the application of traditional merging during the visual encoder due to dimensional incompatibility and spatial structure disruption. While dimensionality reduction can partially mitigate computational costs, significant resource overhead and latency persists, as discussed in Section 3.1~\ref{sec:3_1}. To address this, we design OIM to perform an effective iterative merging that is orthogonal in spatial axes, as illustrated in Fig.~\ref{fig:overview}.

We define visual token sequence $X_v\in\mathbf{R}^{L\times D}$, where $L=H*W$, $L$ is the sequence length, and H and W are the horizontal width and vertical height, respectively. Unlike ToMe which performs similarity computations in one-dimensional space,  OIM iteratively merges tokens in spatial axes for more efficient processing. Initially, we conduct a row merge operation, altering the arrangement of the original tokens to obtain $X'_v\in\mathbf{R}^{H\times W\times D}$. After similarity calculations and binary matching, we acquire $X^1_v\in\mathbf{R}^{H\times(W-m)\times D}$, $m$ is the number of tokens merged per operation. Subsequently, we proceed with the column merge operation, similarly perform parallel calculations in the row dimension to obtain $X^2_v\in\mathbf{R}^{(H-m)\times(W-m)\times D}$. Consequently, OIM achieves true 2D token merging that preserves the structural integrity of visual feature spaces while significantly improving computational parallelism, resulting in hardware-efficient processing.

Additionally, to explicitly differentiate the information density of various tokens and allocate varying levels of attention accordingly, we have adopted a value enhancement strategy. This approach mitigates reliance on attention weights, adapts to FlashA	ttention, and significantly enhances the handling of token quantities, which is formulated as Formula~\ref{equa:equa_3}:

\begin{equation}
	\label{equa:equa_2}
	A=softmax(\frac{QK^T}{\sqrt{d}}) \\
\end{equation}

\begin{equation}
	\label{equa:equa_3}
	O = A(V+log(s))
\end{equation}
where $s$ is a row vector containing the number of the merged token.

\subsection{Spectrum Pruning Unit}
\label{sec:3_3}
The attention-based pruning strategy is incompatible with FlashAttention and exhibits substantial approximation errors~\cite{we_right}, as positional biases systematically distort the evaluation of intrinsic information content of individual tokens. Furthermore, similarity-based methods that focus solely on token-wise proximity often disrupt class balance and the original arrangement of visual tokens. As theoretically analyzed in Section 3.1~\ref{sec:3_1}, low-frequency visual tokens capture more extensive global and cross-modal information, which is preferentially processed by the Transformer architecture. Additionly,  frequency-domain transformation exhibits lower complexity and remains compatible with FlashAttention. Based on this, we design the spectrum pruning unit to progressively prune high-frequency noise tokens during the LLM. We define visual token $X_b[n]\in\mathbf{R}^{D}$, $n$ represents the index of the visual token. We map $X_b[n]$ to the frequency domain through FFT to analyze their spectrum characteristics through Formula~\ref{equa:equa_4}.

\begin{equation}
	\label{equa:equa_4}
	X_b[k] = \sum_{n=0}^{N-1}X_b[n]e^{-i2\pi kn/N} \\
\end{equation}

When applying FFT to visual tokens, we employ Hamming Window~(Formula~\ref{equa:equa_5}) to mitigate frequency leakage artifacts (e.g., spurious high-frequency components induced by row-column transformations). This spectrum smoothing technique progressively attenuates frequency features, suppressing abrupt transitions to yield more accurate spectrum representations that better approximate the true frequency distribution. We obtain a low-pass filter $M_b[k]$ through Formulate~\ref{equa:equa_6} to remove high-frequency noise tokens.

\begin{equation}
	\label{equa:equa_5}
	\begin{split}
		&M_b[k]=\begin{cases}
			0.54-0.46cos(\frac{2\pi n}{N-1}),& \text{ $k \leq \sigma_t$ } \\
			0,& \text{ $ k > \sigma_t $ }
		\end{cases}\\
	\end{split}
\end{equation}

\begin{equation}
	\label{equa:equa_6}
	X'_b[k] = X_b[k]M_b[k]    
\end{equation}

\begin{table*}[htbp]
	\centering
	\scriptsize
	\setlength{\tabcolsep}{2.8mm}
		\caption{Comparative of LUVC with FastV, VTW, and PACT. \textbf{Red. Ratio} denotes the average pruning ratio, \textbf{Latency} represents the prefill latency of the LLM, and \textbf{Speedup} indicates the acceleration ratio. Herein, \textbf{*} signifies that there is a significant discrepancy between the accuracy of the baseline model we reproduce. The best performance is highlighted in \textbf{bold}.}
		\vspace{-0.8em}
		\begin{tabular}{c|c|c|c|cccccc|c}
			\toprule
			Method            & Red. Ratio & Latency &Speedup & VideoMME & MVBench & NextQA & SeedBench & MLVU  & LongVideo & Avg.  \\
			\midrule
			LlavaOV-0.5B      &  0.00\%    & 0.0772  &   --   &  44      &	45.5  & 57.2   & 44.2      & 50.3  & 45.80     & 47.83 \\
			LlavaOV-0.5B*     &  0.00\%    & 0.0772  &   --   &  43.74   &   47.1  & 57.13  & 44.16     & 44.13 & 47.42     & 47.28 \\
			+ FastV        	  &  41.67\%   & 0.0711  &  1.00$\times$  &  43.07   &   46.1  & 56.32  & 43.55     & 43.72 & 45.32     & 46.35 \\
			+ VTW             &  54.17\%   & 0.0596 &  1.30$\times$  &  41.37	&  44.88  &	53.84  & 43.86     & 41.24 & 44.35     & 44.92 \\
			+ PACT            &  48.05\%   & 0.0642  &  1.20$\times$   &  42.7	&  45.17  & 55.16  & 44.8	   & 41.76 & 45.92	   & 45.92 \\
			\textbf{+ LUVC~(ours)}    & 58.46\%  & 0.0610  &  1.27$\times$   &  \textbf{44.37}   &  \textbf{46.85 } & \textbf{57.33 } &\textbf{ 44.96}     & \textbf{44.88} & \textbf{46.90}     & \textbf{47.55 }\\
			\midrule
			LlavaOV-7B	      &  0.00\%    & 0.5194  &  --    & 58.2     & 56.7    &	79.4   & 56.9      & 64.7  & 56.4      & 62.05 \\
			+FastV            &  42.86\%     & 0.2693  &  1.93$\times$   & 57.56	& 57.38	  & 78.08  & 56.72	   & 60.87 & 53.4      & 60.67 \\
			+VTW              &  57.14\%     & 0.2524  &  2.06$\times$   & 46.81	& 45	   &65.43  & 48.44     & 50.36 & 46.97     & 50.50 \\
			+PACT             &  55.88\%     & 0.2711  & 1.92$\times$    & 57.6     &  56.98  & \textbf{78.74} &  56.8     & \textbf{64.7}  & 55.8      & 61.77 \\
			\textbf{+ LUVC~(ours)}    & 62.32\%   & 0.2583  &  2.01$\times$   & \textbf{58.78}    &  \textbf{57.88}  & 78.47  &  \textbf{57.36}    & 63.77 &\textbf{ 56.92}     & \textbf{62.20}\\
			\midrule
			InternVL-2.5-8B	  &    0.00\%    & 1.2480  & --     & 64.20    & 72      & --     & --        & 68.90 & 60.00     & --    \\
			InternVL-2.5-8B*  &    0.00\%   & 1.2480  & --     & 64.50    & 73.25   & 83.43  & 66.41     & 71.80 & 62.30     & 70.28 \\
			+ VTW             &   56.25\%   & 0.6482  &1.93$\times$    & 62.81	& 70.8	  & 80.99  & 65.45	   & 69.32 & 60.36	   & 68.29 \\
			+ PACT            &   55.41\%  & 0.6949  &1.80$\times$     & 63.48	& 71.35	  & 82.3   & 65.18	   & 70.15 &  60.43    & 68.82 \\
			\textbf{+ LUVC~(ours)}   &  62.12\%  & 0.6293 & 1.98$\times$    & \textbf{64.56}    &\textbf{ 71.73}  & \textbf{82.74}  &\textbf{ 65.85}     & \textbf{72.49} & \textbf{62.45}     & \textbf{69.97} \\
			\midrule
			InternVL-2.5-26B  &   0.00\%    & 2.9142  & --     & 66.9    & 75.2    & --     & --        & 72.3  &  59.9     & --    \\
			InternVL-2.5-26B* &   0.00\%    & 2.9142  & --     & 66.9	& 76.47	  & 86.24  & 71.41     & 75.11 &  62.75    & 73.15 \\
			+ VTW             &   66.67\%   & 1.2586  & 2.32$\times$    & 64.56	& 72.43	  & 84.46  & 70.96     &73	   &  61.18	   & 71.10 \\
			+PACT             &    59.61\%    & 1.3548  & 2.15$\times$    & 65.25	& 75.3	  & 85.52  & 70.24     & 73.96 & 60.43     & 71.78 \\
			\textbf{+ LUVC~(ours)}    & 61.32\%      & 1.2255  & 2.38$\times$    & \textbf{66.63}	& \textbf{76.13}	  & \textbf{85.63}  & \textbf{71.28}	   & \textbf{75.23} & \textbf{61.93}	   & \textbf{72.81} \\
			\bottomrule
		\end{tabular}%
	
	\label{tab:Tab1}%
\end{table*}%

After smoothing with the Hamming Window and processing through a low-pass filter, LUVC obtained a more informative low-frequency visual component $X'_b[k]$. Subsequently, we transform $X'_b[k]$ into the original domain to obtain $X'_b[n]$ through the IFFT(Formulate~\ref{equa:equa_7}). Furthermore, we conduct an energy analysis on $X'_b[k]$ by Formalate~\ref{equa:equa_8}, specifically calculating the L2 norm of  $X'_b[k]$, which effectively removes high-frequency noise tokens and returns low-frequency visual tokens with high information content. 

\begin{equation}
	\label{equa:equa_7}
	X'_b[n] = \frac{1}{N}\sum_{k=0}^{N-1}X'_b[k]e^{i2\pi kn/N} \\   
\end{equation}

\begin{equation}
	\label{equa:equa_8}
	E_b[n] = ||X'_b[n]||_2
\end{equation}

By employing this spectrum analysis method for pruning, we effectively eliminate redundant noise tokens and preserve important region, as illustrated in Fig.~\ref{fig:spectrum_visual}, thereby reducing memory usage and computational cost. Simultaneously, this approach provides more focused visual information for cross-modal interactions and aligns well with FlashAttention. Leveraging the feasibility of pruning all visual tokens in the final stages of LLMs as demonstrated by VTW~\cite{VTW}, we integrate the spectrum pruning strategy into the cascade structure. This enables progressive attenuation of visual tokens in the LLM's final stages, accomplishing complete elimination through a controlled, gradual process.
\section{Experiments}

\subsection{Evaluation Datasets}
We evaluate LUVC across diverse benchmarks encompassing video, single-image, and multi-image. Specifically, we conduct performance comparisons on challenging pruning-resistant domains including document and chart comprehension. \textbf{Video Understanding.} For Video tasks, we evaluate VideoMME~\cite{video_mme}, MVBench~\cite{mvbench}, NextQA~\cite{nextqa}, SeedBnech~\cite{seedbench}, MLVU~\cite{mlvu} and LongVideoBench~\cite{long_video_bench}, where VideoMME, MLVU and LongVideoBench are long video benchmarks. \textbf{Single-Image Understanding.} We evaluate LUVC's comprehensive multimodal capabilities across a range of image understanding benchmarks, including MMB~(ZH/EN)~\cite{MMB}, POPE~\cite{POPE}, MMMU~\cite{MMMU}, MMStart~\cite{MMStar} and AI2D~\cite{AI2D}. \textbf{Multi-Image Understanding.} We evaluate LUVC in multi-image perception and understanding across various multi-image benchmarks, including MuirBench~\cite{MuirBench}, BLINK~\cite{BLINK} and RealWorldQA~\cite{realworldqa}.
\textbf{Chart and Document Understanding.} Due to compact spatial layouts and dense textual information, chart and document understanding pose challenging tasks for token compression, we evaluate ChartQA~\cite{chartqa}, DocVQA~\cite{docvqa}, InfoVQA~\cite{infovqa} and TextVQA~\cite{textvqa}. \textbf{Image Caption.} The caption task requires the preservation of more semantic information and exhibits lower tolerance for errors in token compression algorithms. We evaluate the caption capability of LUVC on the Flickr30K~\cite{flickr}.

\begin{table*}[htbp]
	\centering
	\scriptsize
	\setlength{\tabcolsep}{1.3mm}
		\caption{Reault of LUVC and other compress strategies for image benchmarks. The best performance is highlighted in \textbf{bold}.}
		\vspace{-0.8em}
		\begin{tabular}{c|c|c|c|cccccc|ccc|c}
			\toprule
			\multirow{2}[1]{*}{Method}&\multirow{2}[1]{*}{ Red. Ratio}&\multirow{2}[1]{*}{Latency}&\multirow{2}[1]{*}{Speedup}&\multicolumn{6}{c|}{Single Image}&\multicolumn{3}{c|}{Multi Image}&\multirow{2}[1]{*}{Avg.}\\
			\cline{5-13}
			&      &      &  &  MMB(EN)  &   MMB(ZH)   &  POPE  &  MMStar    &MMMUVal  &  AI2D  & MuirBench & BLINK & RealWorld &       \\
			\midrule
			LlavaOV-0.5B	   & 0.00\%  & 0.0462 & --  & 52.1     &   --        &  --    &   37.5     &   31.4   & 57.1   &   25.5    & --    &  55.6     &       \\
			LlavaOV-0.5B*	   &  0.00\%  & 0.0462 & --  & 51.89    &   45.7      &  88.32 &   38.4     &  32.33   & 57.1   &    25.5   &  --   &  55.42    & 49.33 \\
			+FastV             & 41.67\% & 0.0498 & 0.93$\times$ & 51.55   &  43.21	     & 87.09  &	\textbf{39.41}      & 32.33	  & \textbf{56.64}  &   \textbf{26.69}   &  --   &  54.51    & 48.93 \\
			+VTW               &  54.17\% & 0.0398 &1.16$\times$   &49.4	   & \textbf{45.1}	     & 72.56  & 38.11	   & 31.56	  & 55.18  &    25.58  &  --   &  47.71    & 45.65 \\
			+PACT              & 48.05\% & 0.0503 & 0.92$\times$ & 50.6	   & 43.04	     &  86.06 & 39.06	   & \textbf{32.78}	  & 55.34  &    25.62  &  --   & 53.73     & 48.28 \\
			\textbf{+LUVC~(ours)}  & 49.55\% & 0.0470 & 0.98$\times$ & \textbf{51.72}   &	44.59	 & \textbf{88.04} &	39.13	   &  32.56	  &  56.22 &    24.88  &  --   & \textbf{56.47}     & \textbf{49.20} \\
			\midrule
			LlavaOV-7B	       & 0.00\% & 0.1927 & --  & 80.8    &  80.4       & 89.17  &  61.7      &  48.8    & 81.4   &   41.8    &  --   &  66.3     & 68.90 \\
			+FastV             & 42.86\% & 0.1196 & 1.61$\times$ &  79.21   &	79.64	 & 88.83  & 58.56	   & 48.67	  & 79.53  &   41.12   &  --   &  64.18    & 67.47 \\
			+VTW               & 57.14\%& 0.1023 & 1.88$\times$ &  58.59   & 57.99	     & 50.97  & 39.34	   & 43.78	  & 69.43  &   34.85   &  --   &  41.7     & 49.58 \\
			+PACT              & 55.88\% & 0.1096 & 1.76$\times$ & 80.3	   & 79.98	     & 88.93  & 59.79	   &  \textbf{48}	  & 80.86  &   \textbf{43.1}    &  --   &  64.71    & 68.21 \\
			\textbf{+LUVC~(ours)}  & 53.97\% & 0.1205 & 1.60$\times$ & \textbf{80.58}	   & \textbf{80.58}	     & \textbf{89.39}  & \textbf{62.24}	   & 47.89	  & \textbf{81.54}  &   42.46   &  --   &  \textbf{65.62}    & \textbf{68.79} \\
			\midrule
			InternVL-2.5-8B	   & 0.00\% & 0.2985 &  -- & 84.6	   &  82.6       & 90.6	  &   62.8     &	56    & 84.5	 &   --      & 54.8	 &  70.1     & 73.25 \\
			InternVL-2.5-8B*   & 0.00\% & 0.2985 & --  & 83.85   &  83.59      & 88.91  &	  62.93	   & 53.44	  & 84.68  &	 --      & 54.6	 & 70.45     & 72.81 \\
			+VTW               & 56.25\% & 0.1369 & 2.18$\times$  & 83.08   &  81.44	     & 86.09  &   57.33	   & 52	      & 82.84	 &   --      & 52.44 & 63.66     & 69.86 \\
			+PACT              & 55.41\% & 0.1721 &1.73$\times$  & 83.5	   &  82.39	     & 88.21  &   61.06	   & 53.11	  & 83.48	 &   --      & 52.5	 & 67.45     & 71.46 \\
			\textbf{+LUVC~(ours)}  & 56.89\% & 0.1479 & 2.02$\times$ & \textbf{83.85}   &  \textbf{83.25}	     & \textbf{89.19}  &   \textbf{61.4}	   & \textbf{53.22}	  & \textbf{84.1}	 &   --      & \textbf{54.87} & \textbf{69.41}	 & \textbf{72.41} \\
			\midrule
			InternVL-2.5-26B   & 0.00\% & 0.7072 & --  & 85.4	   & 85.5	     & 90.6	  &   66.5	   & 60	      & 86.4	 &   --      & 61.8	 & 74.5	     & 76.34 \\
			InternVL-2.5-26B*  & 0.00\%     & 0.7072 & --  & 86.08   & 85.14	     & 90.3	  &   66.53	   & 53.44	  & 86.46	 &   --      & 61.91 & 75.95	 & 75.73 \\
			+VTW               & 66.67\% & 0.3150 &2.25$\times$  & 84.88   & 83.76	     & 90.09  &   59.73	   & 52.67	  & 84.88	 &   --      & 59.02 & 70.07	 & 73.14 \\
			+PACT              & 59.61\% & 0.3544 &2.00$\times$  & 84.7	   & \textbf{84.45}	     & 90.12  &   62.46	   & \textbf{53.11}	  & 85.36	 &   --      & 59.23 & 72.03	 & 73.93 \\
			\textbf{+LUVC~(ours)}  & 55.1\%& 0.3379 & 2.09$\times$ & \textbf{84.97}   & \textbf{84.45}	     & \textbf{90.32}  &   \textbf{65.2}	   & 53	      & \textbf{85.72}	 &   --      & \textbf{61.75} & \textbf{74.9}	     & \textbf{75.04} \\
			\bottomrule
		\end{tabular}%
	
	\vspace{-1.8em}
	\label{tab:Tab2}%
\end{table*}%

\subsection{Evaluation Setup}
To comprehensively validate the applicability of LUVC across different VLM architectures and parameter scales, we conducted evaluations on all benchmarks using four distinct models: LLaVA-OV-0.5B, LLaVA-OV-7B, InternVL-2.5-8B, and InternVL-2.5-26B. To rigorously demonstrate the algorithmic advantages of LUVC, we selected three representative baselines for comparison: FastV (a classical attention-score-based method), PACT (the state-of-the-art similarity clustering algorithm), and VTW (an ultimate pruning-based approach). For fair comparison, we carefully tuned the parameters of each algorithm to achieve optimal performance on VLMs while maintaining comparable acceleration ratios. In addition to standard benchmark metrics, we measured the prefill-phase latency and average pruning rate~($Red. Ratio=\frac{L_p}{L_o}$) for each algorithm, where $L_p$ is the average visual token sequences after compression and $L_o$ is the average visual token sequences without compression. Notably, we adhered to the original hyperparameter configurations in paper for LLaVA-OV-7B and InternVL2.5-8B, while performing additional tuning (as no official settings were provided) to optimize performance on LLaVA-OV-0.5B and InternVL2.5-26B.

For LUVC, regarding the video benchmark, we merge approximately 35\%-40\% of the tokens during the visual encoder. For single-image or multi-image benchmarks, we merge around 15\%-20\% of the tokens during the visual encoder. For chart and document comprehension benchmarks, to maintain sufficient information and spatial layout, we do not perform OIM. \textbf{Additional experimental details and results on QwenVL2.5 will be provided in the supplementary material.}

\subsection{Main Results}
\textbf{Video Understanding.} We conducted a comparative evaluation of LUVC, FastV, VTW, and PACT based on the LLaVA-OV series. Notably, since FlashAttention was enabled during the testing of InternVL2.5, we do not compare FastV on InternVL2.5. As shown in Table~\ref{tab:Tab1}, LUVC demonstrates superior video understanding performance compared to other pruning algorithms while maintaining a similar speedup. Specifically, LUVC achieves lossless performance across all LLaVA-OV series models. For the InternVL2.5 series, it incurs only a marginal accuracy degradation of approximately 0.3\%, substantially outperforming the suboptimal PACT algorithm. Taking InternVL2.5-8B as an example, LUVC exhibits average video understanding metrics that surpass VTW and PACT by 1.68\% and 1.15\%, respectively.

\textbf{Single-Image Understanding} and \textbf{Multi-Images Understanding.} For both single-image and multi-image benchmarks, LUVC maintains its performance advantage, achieving state-of-the-art results across evaluations on the LLaVA-OV and InternVL2.5, as shown in Table~\ref{tab:Tab2}. Taking InternVL2.5-8B as an example, LUVC surpasses VTW and PACT by 2.55\% and 0.95\%, respectively, in average video metrics. Notably, in the multi-image evaluation of the InternVL2.5 series, LUVC significantly outperforms the suboptimal PACT algorithm by over 2\%. These results further validate the generalizability of the LUVC algorithm across different evaluation tasks and model architectures.


\begin{table}
	\scriptsize
	\setlength{\tabcolsep}{0.3mm}
	\label{tab:freq}
 \caption{Comparison between LUVC and other token compression methods in chart and document tasks.}
	\begin{tabular}{c|c|c|ccccc}
		\toprule
		Method            &  Latency   & Speedup & ChartQA & DocVQA   & InfoVQA & TextVQA &  Avg.    \\
		\midrule
		InternVL-2.5-8B	  &  0.2985    & --      &  84.8   &  --      &   --    &   79.1  &  --      \\
		InternVL-2.5-8B*  &  0.2985    & --      &  83.36  & 91.96    &   75.45 &   78.99 &  82.44   \\
		+ VTW             &  0.1804    & 1.65    &  76.08  & 80.59    &  67.94  &   60.83 &  71.36   \\
		+ PACT            &  0.1980    & 1.51    &  75.96  &  84.32   &   64.67 &   72.73 &  74.72    \\
		\textbf{+ LUVC~(ours)}    &  0.1857    & 1.61    &  \textbf{82.44}  &  \textbf{91.48}   &   \textbf{74.75} &   \textbf{77.80} &   \textbf{81.62} \\
		\midrule
		InternVL-2.5-26B  &  0.7454    & --      &  87.2   &  --      &  --     &   82.4  &    --    \\
		InternVL-2.5-26B* &  0.7454    & --      &  85.4   &  93.78   &   78.98 &  83.26  &   85.36  \\
		+ VTW             &  0.4637    & 1.61    &  82.24  &  88.78   &   75.52 &  73.33  &   79.97  \\
		+PACT             &  0.4690    & 1.59    &  76.36  &  85.79   &   68.46 &  78.56  &   77.29  \\
		\textbf{+ LUVC~(ours)}    &  0.4302    & 1.73    &  \textbf{85.16}  &  \textbf{92.70}   &   \textbf{77.55} &  \textbf{80.03}  & \textbf{83.86} \\
		\bottomrule
	\end{tabular}
	
	\label{tab:chart_doc_eval}%
	\vspace{-0.9em}
\end{table}

\subsection{Extra Experiments for Chart and Document Tasks}
Chart and document comprehension tasks impose stringent demands on localized details and spatial layouts in images, presenting challenges for compression algorithms. To assess the efficacy of LUVC, we conduct benchmark evaluations using the InternVL2.5 model series on four challenging datasets: ChartQA, DocVQA, InfoVQA, and TextVQA. Due to task complexity, we sacrifice partial acceleration benefits during evaluation. Specifically, we disable the OIM in LUVC and substantially delay the pruning layers in both PACT and VTW to prevent performance collapse. As illustrated in Table~\ref{tab:chart_doc_eval}, under similar latency, PACT and VTW still exhibit unacceptable performance degradation, whereas LUVC maintains a stronger performance advantage.

\begin{table}
	\scriptsize
 \caption{Comparison between LUVC and other token compression methods in caption tasks Flickr30K~\cite{flickr}.}
	\setlength{\tabcolsep}{0.4mm}
	\begin{tabular}{c|c|cccccc}
		\toprule
		Method            & Speedup & BLEU-1 & BLEU-2   & METEOR & ROUGE-L &  CIDEr & SPICE    \\
		\midrule
		InternVL-2.5-8B	  & --      & 0.248     &   0.156      &  0.139     &  0.329   &   0.669  & 0.236   \\
		+ VTW             & 1.65    & 0.203    & 0.119    &  0.115   &  0.291   &  0.448  &0.191  \\
		+ PACT            & 1.51    &  0.231   &  0.144   &  0.133   &  0.32   &   0.602  & 0.23 \\
		\textbf{+ LUVC~(ours)}    & 1.61    &  \textbf{0.243}  &  \textbf{0.154}   &   \textbf{0.137} &   \textbf{0.33} &   \textbf{0.666}& \textbf{0.235}\\
		\midrule
		InternVL-2.5-26B  & --      & 0.255  &  0.159     &  0.140   &  0.328  &   0.666   & 0.234\\
		+ VTW             & 1.61    & 0.242 & 0.149 & 0.132 & 0.321 & 0.601  & 0.224 \\
		+PACT             & 1.59    & 0.24  & 0.149  &  0.134 &0.32 &  0.63  &0.229\\
		\textbf{+ LUVC~(ours)}    & 1.73    &  \textbf{0.25}  &  \textbf{0.156}   &   \textbf{0.138} &  \textbf{0.326}  & \textbf{0.664}&\textbf{0.232}\\
		\bottomrule
	\end{tabular}
	
	\label{tab:caption_eval}%
	\vspace{-0.9em}
\end{table}

\subsection{Extra Experiments for Image Captioning Task}
The captioning capability provides a critical measure of whether key visual semantics are effectively preserved. To further demonstrate the generalization ability of LUVC, we conducted additional experiments on the Flickr~\cite{flickr} captioning task. The experimental results are summarized in the Table~\ref{tab:caption_eval}, where a comparative analysis was performed based on InternVL2.5—8B/26B. LUVC exhibits almost no performance degradation, whereas both PACT and VTW show significant declines. Specifically, PACT exhibits a 6.7 PCT decrease in CIDEr score, while VTW shows a 22 PCT reduction with InternVL2.5-8B, indicating diminished content relevance. The observed decrease in BLEU and ROUGE scores corresponds to a substantial deterioration in caption coherence and quality.
In contrast, LUVC exhibits only minor reductions of 0.5 and 0.2 in BLEU-1 and BLEU-2 scores, while its CIDEr score surpasses that of PACT by 6.4 percentage points and VTW by 21.7 points. The same experimental conclusion is supported by the results from InternVL2.5-26B. These findings demonstrate that LUVC effectively preserves key semantic information while maintaining a comparable acceleration ratio, thereby sustaining superior model captioning performance—an advantage not achieved by other training-free compression algorithms.

\begin{figure}[htbp]
	\centering
	\begin{subfigure}{0.15\textwidth}
		\includegraphics[width=\linewidth]{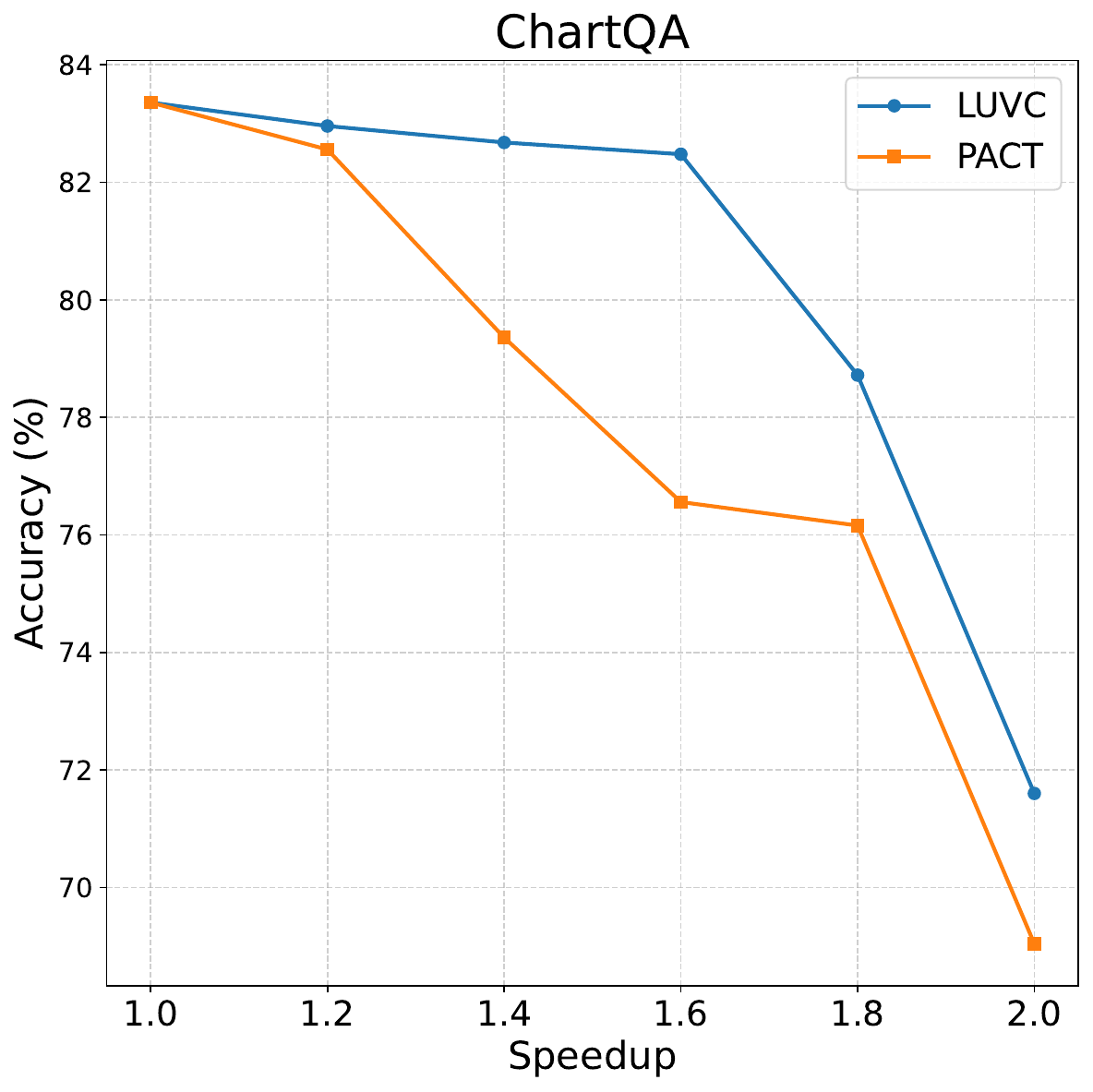}
		\caption{ChartQA}
		\label{fig:chartqa_speedup}
	\end{subfigure}
	\hfill 
	\begin{subfigure}{0.15\textwidth}
		\includegraphics[width=\linewidth]{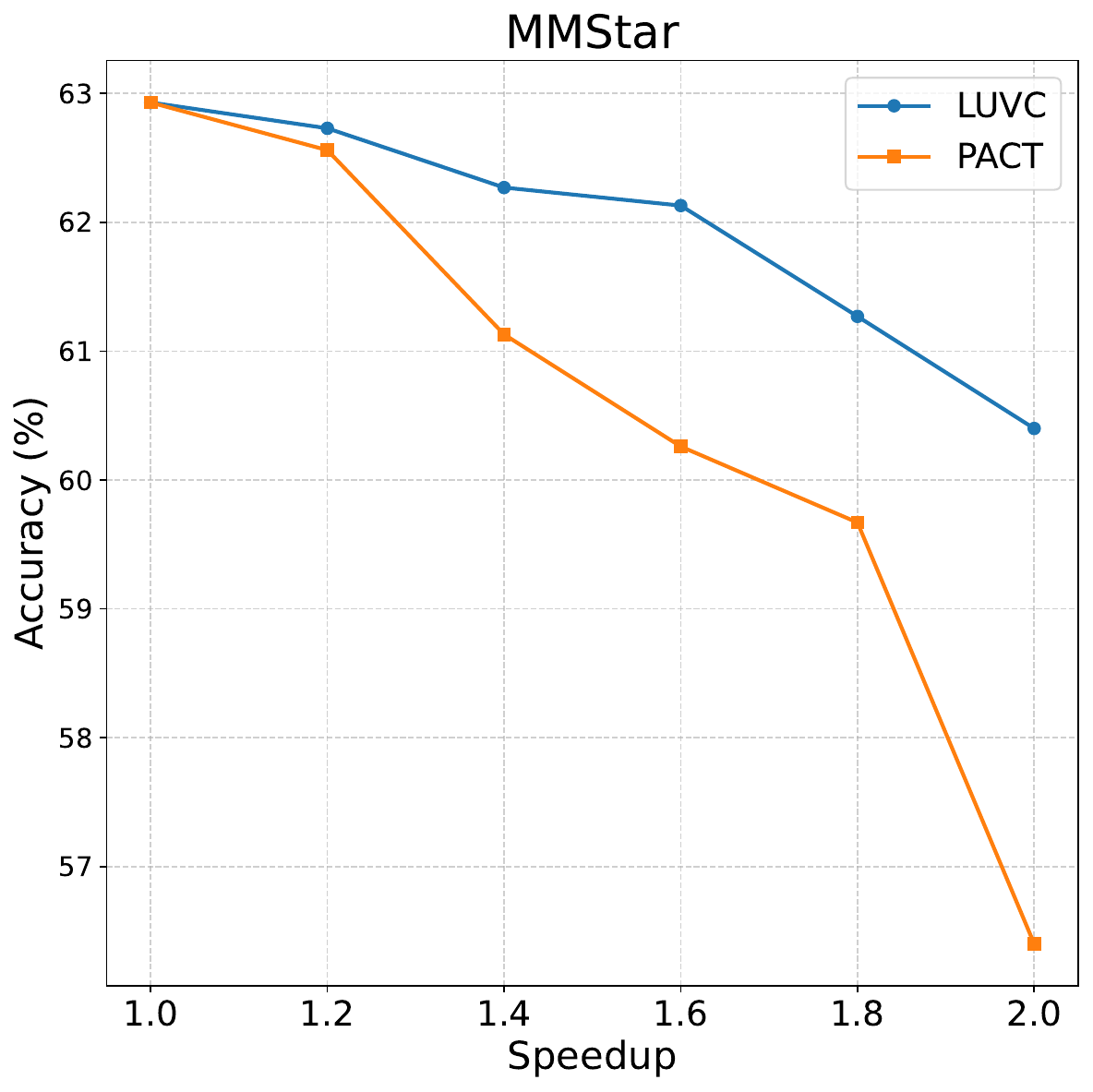}
		\caption{MMStar}
		\label{fig:mmstar_speedup}
	\end{subfigure}
	\hfill 
	\begin{subfigure}{0.15\textwidth}
		\includegraphics[width=\linewidth]{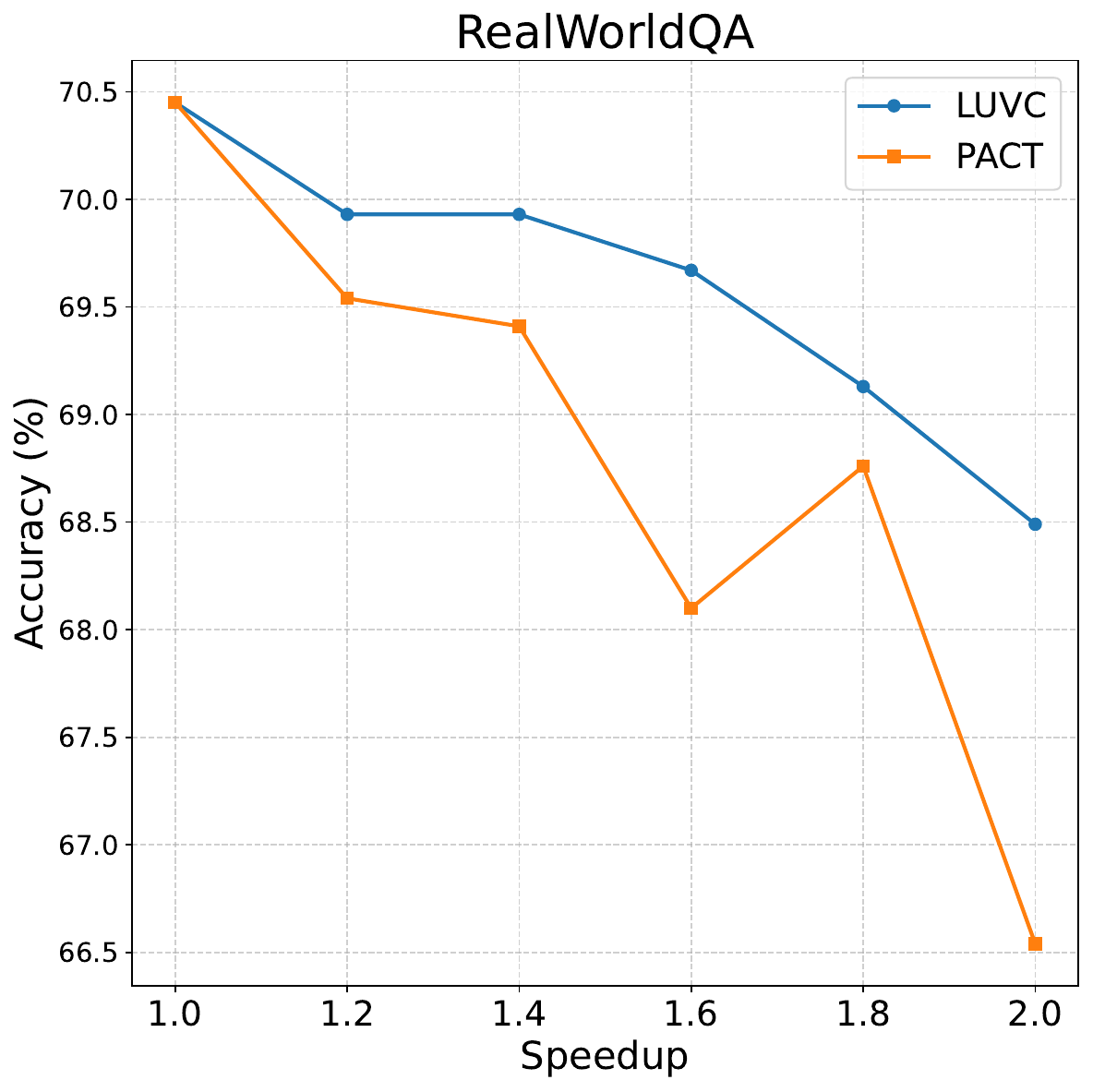}
		\caption{RealWorldQA}
		\label{fig:realworldqa_speedup}
	\end{subfigure}
	\caption{Performance comparison of LUVC and PACT under different \textbf{Speedup}.}
	\label{fig:ablation_speedup}
	\vspace{-1.5em}
\end{figure}

\begin{figure}[htbp]
	\centering
	\includegraphics[width=6.2cm]{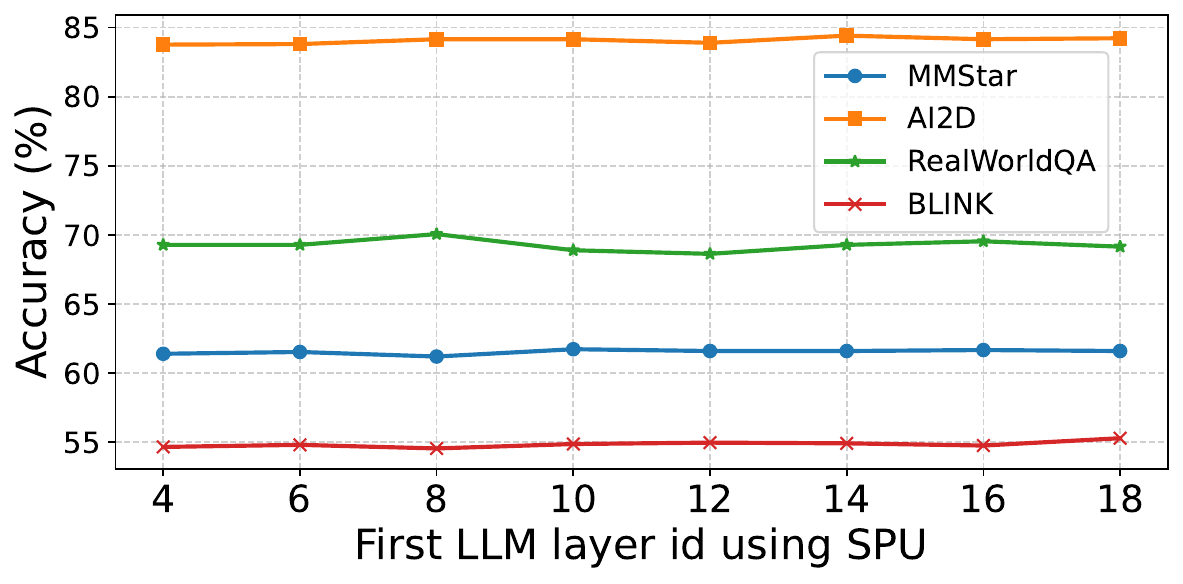}
	\caption{Performance Comparison of LUVC using SPU from different LLM layers.}
	\label{fig:ablation_llm_start}
	\vspace{-1.0em}
\end{figure}

\subsection{Throughput ratio, GPU memory and Algorithm latency}
As demonstrated in Table~\ref{tab:gpu_memory}, we control comparable performance between PACT/VTW and LUVC by either postponing the initial pruning layer or adjusting pruning thresholds. Under this condition, LUVC attains a Throughput Ratios of 214\%, significantly surpassing both PACT and VTW, while simultaneously maintaining the lowest GPU memory.

To provide a more intuitive comparison of the computational latency introduced by different algorithms during model inference, we conducted a comparative analysis of PACT, VTW, and our proposed LUVC. As shown in Table~\ref{tab:algo_latency}, both the algorithm-specific overhead and the overall model inference latency are presented, with results averaged over multiple test runs using an example image with 12 patches. It can be observed that although LUVC employs an iterative pruning strategy, it does not introduce additional latency burden to the model inference process.

\begin{table}
	\scriptsize
	
	\setlength{\tabcolsep}{1.8mm}
	\caption{Throughput Ratios, Red. Ratio, and GPU usage of different methods in InternVL2.5-8B.}
	\label{tab:freq2}
	\begin{tabular}{c|cccc}
		\toprule
		Method            &  No reduction   &  VTW  & PACT   & \textbf{LUVC~(ours)} \\
		\midrule
		Red. Ratio	      &   0.00\%        & 40.25\%  & 42.00\%       &  \textbf{56.89}\%     \\
		Throughput Ratio  &  100\%          & 140\% & 155\%  &   \textbf{214\%}           \\
		GPU Memory~(GB)   &  21.79          & 19.54 & 19.68  &   \textbf{18.68}        \\
		\bottomrule
	\end{tabular}
	
	\vspace{-0.6em}
	\label{tab:gpu_memory}%
	
\end{table}

\begin{table}
	\scriptsize
	
	\setlength{\tabcolsep}{2.8mm}
	\caption{Algorithm and Model Latency~(senonds).}
	\label{tab:freq3}
	\begin{tabular}{c|cccc}
		\toprule
		Method               &  VTW  & PACT   & \textbf{LUVC} & The whole model\\
		\midrule       
	InternVL2.5-8B	   &       0.011   & 0.033 &  0.031  &   0.2985        \\
	InternVL2.5-26B	   &        0.019     & 0.062    &  0.049& 0.7454        \\
		\bottomrule
	\end{tabular}
	
	\vspace{-0.6em}
	\label{tab:algo_latency}%
	
\end{table}

\subsection{Ablation Studies}

\textbf{The design of core comments of LUVC.} First, we compare the performance impact of OIM and SPU using different settings, as shown in Table~\ref{tab:ablation_core}. The experimental results demonstrate that OIM and SPU jointly contribute to the compression efficacy and accuracy maintenance of LUVC. Moreover, value enhancement~(\textbf{VE}) and Hamming Window~(\textbf{HW}) optimization positively influence OIM and SPU, respectively.

\textbf{The effect of OIM.} Orthogonal Iterative Merging performs orthogonal iterative merging of visual tokens in the spatial scale during ViT. This process can be replaced by 2D downsampling or 2D random selection, and we conduct ablation studies on different strategies as shown in the Table~\ref{tab:ablation_oim}. Additionally, we compare the 1D merging strategy of ToMe~\cite{ToMe} to achieve a similar compression ratio. As presented in Table~\ref{tab:ablation_oim}, ToMe fails to preserve the 2D spatial structure, resulting in disrupted spatial information and severely compromising model performance, with an average metric drop of 2.53 pct. Although 2D downsampling maintains spatial structure, it applies uniform downsampling globally without considering the semantic information of tokens, leading to an average performance decrease of 1.22 pct. We also evaluated a 2D random selection strategy, and the results indicate that random pruning in 2D space cause performance degradation. In contrast, OIM achieves the best performance among all compared methods.

\textbf{The effect of Speedup.} Fig.~\ref{fig:ablation_speedup} shows the performance variations of LUVC across ChartQA, RealWorldQA, and MMStar, under different Speedup ratios. As the Speedup increases, the model exhibits a gradual performance decline while maintaining reliable accuracy. For comparison, we also plot the corresponding performance trends of PACT and VTW. The results demonstrate that LUVC consistently outperforms these baselines across all Speedup ratios without experiencing abrupt performance degradation.

\textbf{The effects of initial LLM layer selection for SPU.} As illustrated in Fig.~\ref{fig:ablation_llm_start}, we investigate the impact of pruning starting layer selection on model performance based on InternVL2.5-8B during the LLM pruning. Notably, owing to the low computational complexity of FFT, we can implement cascade progressive pruning without causing computational bottlenecks. Experimental results demonstrate that LUVC maintains stable accuracy as the pruning starting layer shifts forward, in stark contrast to algorithms like VTW, which exhibit catastrophic performance collapse when pruning is applied to early-stage layers. This validates the stability and layer-agnostic robustness of LUVC.

\begin{table}
	\scriptsize
 \caption{Ablation studies on the core components of LUVC.}
	\setlength{\tabcolsep}{0.9mm}
	\begin{tabular}{c|c|c|cccc|c}
		\toprule
		Method                  &  Latency   & Speedup & MMStar & POPE      & BLINK   &    AI2D &  Avg.    \\
		\midrule
		InternVL-2.5-8B*        &  0.2985    &  --     &  62.93 &  88.91    & 54.8    &   84.5  & 72.79   \\
		+ OIM~(wo \textbf{VE})  &  0.2615    & 1.14    &  60.7  &  88.46    & 54.65   &   83.9  & 71.93     \\
		+ OIM~(\textbf{VE})     &  0.2615    & 1.14    &  61.82  &  89.11    & 54.93   &   84.21 & 72.52    \\
		+ SPU~(wo \textbf{HW})  &  0.2075    & 1.44    &  61.51  &  89.04    & 54.66   &  84.10  & 72.33   \\
		+ SPU~(\textbf{HW})     &  0.2075    & 1.44    &  61.83  &  89.31    & 54.65   &  84.23  &  72.51   \\
		\midrule
		\textbf{LUVC}           &  0.1658    & 1.80    &  61.4  & 89.19     & 54.87   &   84.1  &   72.39  \\
		\bottomrule
	\end{tabular}
	
	\label{tab:ablation_core}%
	\vspace{-1.5em}
\end{table}

\begin{table}
	\scriptsize
 \caption{Ablation studies on OIM and other compression methods.}
	\setlength{\tabcolsep}{2.4mm}
	\begin{tabular}{c|cccc|c}
		\toprule
		Method              &    MMStar & POPE      & BLINK   &    AI2D &  Avg.    \\
		\midrule
		InternVL-2.5-8B*         &  62.93 &  88.91    & 54.8    &   84.5  & 72.79   \\
        + ToMe~\cite{ToMe}     & 59.78 & 84.28   & 54.00   &  82.96   & 70.26   \\
        + 2D random Prune        &  60.83  &  87.71    & 54.07   &   83.66 & 71.57    \\
        + nearest interpolate      &  61.05  &  88.71    & 54.24   &   83.43 & 71.86    \\
        + bilinear interpolate      &  61.12  &  88.31    & 53.91   &   83.91 & 71.81    \\
        
		\midrule
		+ OIM~(\textbf{VE})     &   61.82  &  89.11    & 54.93   &   84.21 & 72.52       \\
		\bottomrule
	\end{tabular}
	
	\label{tab:ablation_oim}%
	\vspace{-1.5em}
\end{table}

\section{Conclusion}
We propose an ultimate visual tokens compression framework named LUVC based on spectrum theory and spltial merging. This scheme does not require any training and can be seamlessly integrated into various architectures and parameter quantities of VLMs. LUVC first merges highly similar visual tokens through a orthogonal iterative merge strategy during visual encoding while preserving spatial structural information. Furthermore, LUVC incorporates a dedicated Spectrum Pruning Unit and employs it in a cascaded manner across the LLM. SPU leverages the low-pass filtering property of Transformers to progressively integrate visual information into cross-modal queries by pruning high-frequency redundant tokens. This ultimately achieves complete compression of visual tokens in the LLM. Through its pruning strategy design across all stages, LUVC achieves approximately double the inference speed with minimal accuracy error. Experiments demonstrate that LUVC is applicable to diverse multimodal understanding scenarios, including videos,  single images, and multi-images.

{
    \small
    \bibliographystyle{ieeenat_fullname}
    \bibliography{main}
}

\newpage

\section{Appendix}
\subsection{Experiments on Qwen2-VL-7b-Instruct}

To further validate the generalizability of our proposed LUVC across diverse VLM architectures, we conduct additional experiments on Qwen2-VL-7B-Instruct~\cite{qwen2_vl}. As shown in the Table~\ref{tab:Tab1}, we evaluate six datasets, including MMStar~\cite{MMStar}, AI2D~\cite{AI2D}, RealWorldQA~\cite{realworldqa}, DocVQA~\cite{docvqa}, POPE~\cite{POPE}, and ChartQA~\cite{chartqa}, spanning single-image, multi-image, and dense document understanding tasks. The latency is measured on 100 randomly sampled documents from DocVQA, with mean processing times computed for each algorithm on GPU. The experimental results demonstrate that, under comparable speedup ratios, LUVC achieves an average performance improvement of \textbf{+2.52} over PACT, the state-of-the-art similarity-aware method, while surpassing the attention-aware representative method FastV by \textbf{+3.32}. Additionally, it outperforms VTW by \textbf{+24.45} percentage points, respectively. To date, LUVC has undergone extensive experimental validation across VLM models from the Qwen2VL, InternVL2.5, and LLaVA-OV series. Substantial evidence confirms that the proposed SPU and OIM strategies in LUVC deliver robust and stable performance in accelerating VLM inference. 

\begin{table*}[htbp]
	\centering
	\scriptsize
	\setlength{\tabcolsep}{4.0mm}
		
		\vspace{-0.8em}
		\begin{tabular}{c|c|c|cccccc|c}
			\toprule
			Method                  & Latency &Speedup & MMStar & AI2D & RealWorldQA & DocVQA & ChartQA  & POPE  & Avg.  \\
			\midrule
			Qwen2VL-7B-Instruct	    &  0.3569 &  --    &  56	& 79.9 &	65.1     &	93.9  &   80.8	 & 88.58 &77.38  \\
			+FastV                  &  0.3594 &  0.99$\times$    & 51.5	& 76.2 &	63.81	 & 86.6	  & 75.32	 & 85.76 & 73.20 \\
			+VTW                    &  0.2437 &  1.46$\times$   &  \textbf{ 55.7}	& 77.4 & 64.18	 & 10.84  & 18.12    & 86.19 & 52.07  \\
			+PACT                   &  0.2179 &  1.64$\times$    &  54.8	& 78.4 & 58.95	     & 90.5	  & 76	     & 85.35 & 74  \\
			
			\textbf{+ LUVC~(ours)}  &  0.2335 &  1.53$\times$  &  55.66	& \textbf{78.86}&  \textbf{65.23} & \textbf{91.33}  & \textbf{80.08}	 & \textbf{87.98} &\textbf{76.52}   \\
			
			\bottomrule
		\end{tabular}%
	\caption{Comparative of LUVC with FastV, VTW, and PACT on Qwen2VL-7B. \textbf{Latency} represents the prefill latency of the LLM, and \textbf{Speedup} indicates the acceleration ratio. Herein, the best performance is highlighted in \textbf{bold}.}
	\label{tab:Tab1}%
	\vspace{-0.8em}
\end{table*}%

\subsection{Implementation Details of Other Methods}
\textbf{For PACT~\cite{pact}.} As described in the PACT, the authors conducted validation experiments on the Qwen2-VL-7B, InternVL2-8B, and LLaVA-OV-7B. We reused their original settings and reproduced the publicly reported performance using the open-source code on LLaVA-OV-7B and Qwen2-VL-7B. Given the architectural consistency between InternVL2 and InternVL2.5, we applied the same parameter configurations from InternVL2-8B to InternVL2.5-8B, achieving comparable performance (including publicly reported metrics such as speedup and performance degradation). For LLaVA-OV-0.5B, since the authors did not provide relevant experiments, we adjusted the pruning starting layer. Using the settings from LLaVA-OV-7B led to performance collapse, so we select the 8th layer to achieve stable performance. Similarly, for InternVL2.5-26B, we chose the 11th layer as the starting pruning layer to ensure stable performance.

\textbf{For FastV~\cite{FastV}.}  The core principle of the FastV lies in its text-guided visual pruning mechanism based on text-image attention. The authors operate pruning at the second layer of the LLM. However, given that modern computational architectures have moved away from traditional attention computation in favor of optimized approaches such as FlashAttention~\cite{flash_attn}, we adapt the method for LLaVA-OV and Qwen2VL series by retaining the relevant query and key matrices to recompute the attention required by FastV. To achieve an optimal balance between pruning depth and model performance, we systematically adjusted the selection of pruned layers. Specifically, for LLaVA-OV-0.5B, we retaine the original design proposed by the authors, implementing pruning at the second layer. In contrast, for both LLaVA-OV-7B and Qwen2-VL-7B, we strategically shift the pruning layer to the fourth layer to enhance performance stability. This deliberate layer selection strategy ensures robust performance preservation while maintaining computational efficiency under the constraints of contemporary attention mechanisms.

\textbf{For VTW~\cite{VTW}.} The VTW algorithm posits that visual tokens become increasingly redundant in later layers of LLMs. The researchers employed a meticulously designed pruning layer selection strategy to identify the optimal starting layer for discarding all visual tokens. In our reproduction of their method, we fine-tuned the pruning layer within a window around the originally suggested position to achieve the best balance between performance and inference speed. Specifically, we identified the following optimal pruning layers for each model:
\begin{itemize}
	\item LLaVA-OV-0.5B: Layer 8
	\item LLaVA-OV-7B/Qwen2-VL-8B: Layer 12
	\item InternVL2.5-8B: Layer 14
	\item InternVL2.5-26B: Layer 20
\end{itemize}

This configuration is carefully designed to minimize adverse effects on model performance while preventing potential performance collapse. The layer selection follows a clear scaling pattern correlated with model size, demonstrating that larger-scale models preserve their representational capacity by implementing pruning at later stages.

\subsection{Evaluation Datasets and Metrics}
\begin{algorithm}[tb]
	\caption{Orthogonal Iterative Merging}
	\label{alg:oim}
	\textbf{Input}: $T_v\in R^{n_v\times D}$\\
	\textbf{Parameter}: $r$, the number to merge in each spatial axes \\
	$l_i$, the layer index of visual encoder  \\
	$L_h$, the visual encoder layers to operate merging in height. \\
	$L_w$, the visual encoder layers to operate merging in width. \\
	\begin{algorithmic}[1]
		\FOR {i,j in zip($L_h$,$L_w$)} 
		\STATE assert j=i+1
		\ENDFOR
	\end{algorithmic}
	
	\textbf{Output}: The visual tokens $T_v'$ after Orthogonal Iterative Merging.
	
	\begin{algorithmic}[2] 
		\STATE Define: \\
		\textbf{$MLP_{qkv}$}: The map function of qkv. \\
		\textbf{$FA$}: Flash Attention \\
		\textbf{$s$}: vector containing the number of the merged tokens. \\
		\textbf{$OIM_h$}: Orthogonal Iterative Merging in height. \\
		\textbf{$OIM_w$}: Orthogonal Iterative Merging in width. \\
		
		\IF {$l_i$ in $L_h$}
		\STATE $q,k,v=MLP_{qkv}(T_v)$
		\STATE $T_v'=FA(q,k,v+log(s))$
		\STATE $T_v' = OIM_h(T_v')$ 
		\ELSIF {$l_i$ in $L_w$}
		\STATE $q,k,v=MLP_{qkv}(T_v)$
		\STATE $T_v'=FA(q,k,v+log(s))$
		\STATE $T_v' = OIM_w(T_v')$
		\ELSE
		\STATE continue.
		\ENDIF
	\end{algorithmic}
\end{algorithm}

We conduct extensive evaluations across multiple benchmarks. We summarize their respective evaluation metrics, as detailed in the Table~\ref{tab:metric}. The ANLS (Average Normalized Levenshtein Similarity) metric serves as the standard evaluation criterion in DocVQA adn InfoVQA tasks, quantifying the similarity between model-generated answers and reference answers. It is noteworthy that, to ensure a fair and objective comparison of algorithm performance, we have excluded benchmarks that require third-party model evaluation.

\begin{algorithm}[tb]
	\caption{Spectrum Pruning Unit}
	\label{alg:spu}
	\textbf{Input}: $T_v\in R^{n_v\times D}$\\
	\textbf{Parameter}: $n_v'$, the visual token length to keep \\
	$l_i$, the layer index of LLM \\
	$L_p$, the all LLM layers to operate low-pass filtering \\
	\textbf{Output}: Low-frequence visual tokens $T_v'$.
	\begin{algorithmic}[1] 
		\STATE Define: \\
		\textbf{FFT}: Fast Fourier Transform;  \\
		\textbf{IFFT}: Inverse Fast Fourier Transform. \\
		\textbf{HW}: Hamming Window. \\

		\IF {$l_i$ in $L_p$}
		\STATE $F_v=FFT(T_v)$, get frequence-domain feature of $T_v$
		\STATE $F_v'=HW(F_v)$, get low-frequence feature $F_v'$.
		\STATE $T_v'=IFFT(F_v')$, get low-frequence tokens $T_v'$.
		\STATE $E_v=||(T_v')||_2$, get the energy  $E_v$ of visual tokens.
		
		\STATE $IDX_l$ = $topk(E_v,n_v').index$
		\STATE $T_v'$ = $T_v[IDX_l]$ 
		\ELSE
		\STATE continue.
		\ENDIF
	\end{algorithmic}
\end{algorithm}

\begin{table}[htbp]
	\centering
	\scriptsize
	\setlength{\tabcolsep}{6.0mm}

		\begin{tabular}{c|c|c}
			\toprule
			Benchmark               & Split   & Metric \\
			\midrule
			VideoMME	            & --      &  Acc.    \\
			MVBench                 & --        &  Acc.   \\
			NextQA                  &  MC       &  Acc.    \\
			SeedBench               &  Video       &  Acc.    \\
			MLVU                    &   --      &  Acc.    \\
			LongVideo               &    --     &  Acc.    \\
			\midrule
			MMB      	            &    Val     &  Acc.    \\
			POPE                    &     --    &  F1    \\
			MMStar                  &    --     &  Acc.   \\
			MMMUVal                 &    Val     &  Acc.    \\
			AI2D                    &    --     & Acc.    \\
			MuirBench               &    --     &  Acc.    \\
			BLINK                   &    --     &  Acc.    \\
			RealWorld               &    --     &  Acc.    \\
			\midrule
			ChartQA                 &   Val      &  Relaxed Acc.    \\
			DocVQA                  &   Val      & ANLS    \\
			InfoVQA                 &   Val      &  ANLS    \\
			TextVQA                 &   Val      &  Official metric    \\
			\bottomrule
		\end{tabular}%
	\caption{Dataset Splits, and Evaluation Metrics Used in Our Experiments.}
	\label{tab:metric}%
	\vspace{-0.8em}
\end{table}%

\begin{figure*}
	\begin{center}
		\includegraphics[width=16.6cm]{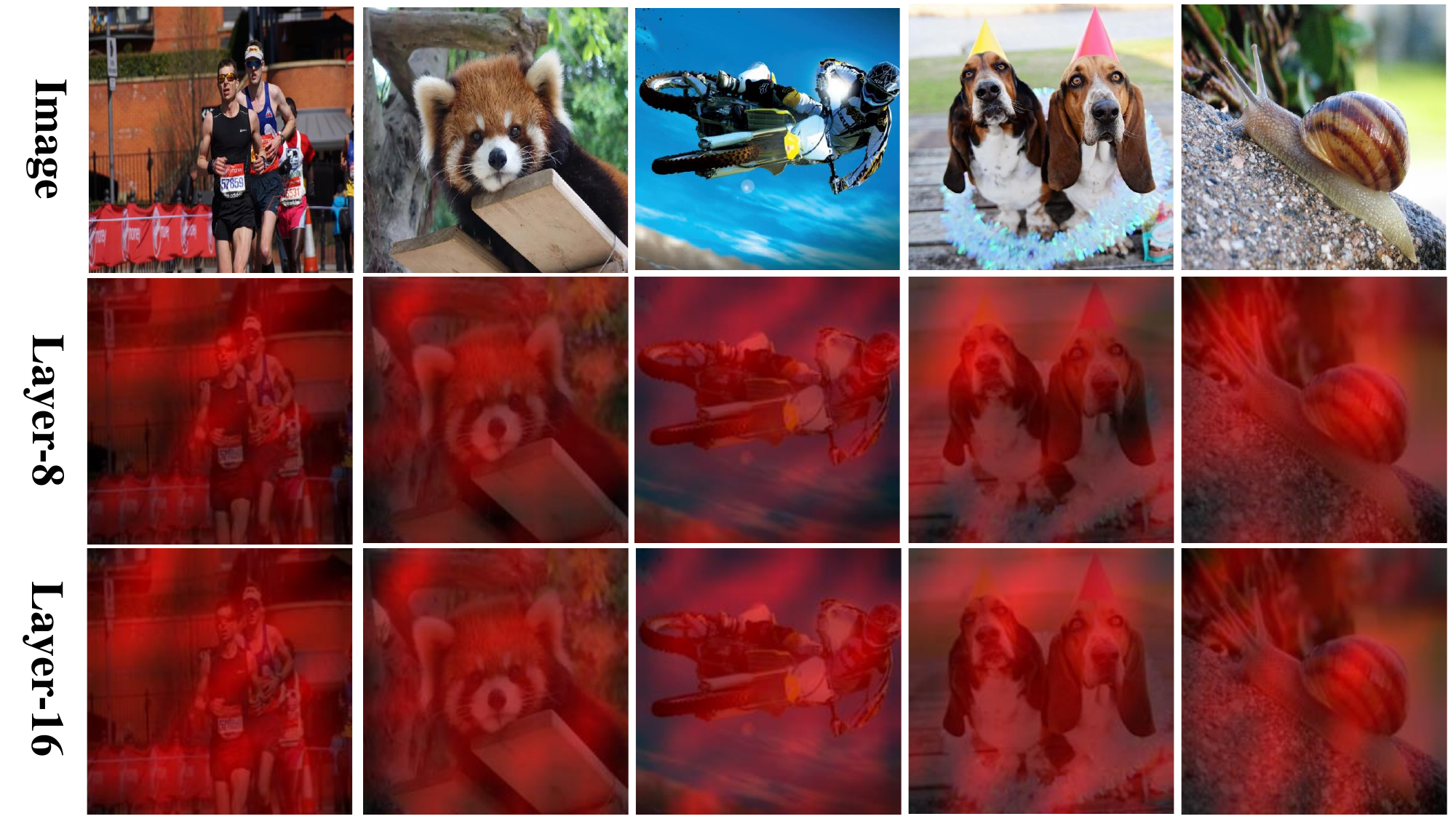}
	\end{center}
	\caption{The visualization results of the mapping relationship between low-frequency visual tokens and the original image. These results were obtained from different LLM layers of InternVL2.5-8B.
	}
	\label{fig:appendix_visual}
\end{figure*}

\subsection{Iterative Strategy and Pseudocode of OIM}

As shown in the Algorithm~\ref{alg:oim}, we demonstrate the detailed process of orthogonal iterative merging~(\textbf{OIM}) execution. For the visual encoder, we opt to merge visually similar tokens at the intermediate stage of the network. We employ an alternating iterative strategy for merging the height and width dimensions, with the merge number configured as 2.  In each OIM in spatial axes, one height-dimension merging and one width-dimension merging are performed. The number of OIM operations is set to 3 for video scenes, while for text-image scenes, it can be adjusted to either 1 or 2. We have established a flexible parameter adjustment strategy that can be controlled according to different acceleration requirements.

Specifically, for the Qwen2VL model, due to its adoption of the anyres strategy, we no longer employ a fixed number of merges. Instead, the merging operations along the spatial dimensions are dynamically adjusted based on the aspect ratio of the input image.

\subsection{Pseudocode of Low-Frequency Filter in SPU}

As illustrated in the Algorithm~\ref{alg:spu}, we provide a detailed demonstration of the spectrum pruning unit~(SPU) execution process. For SPU operations, we configure two key parameters:  the initial pruning layer ($l_0$) and the pruning interval ($l_{\delta}$). Specifically:

\begin{itemize}
	\item LLaVA-0.7/7B and Qwen2VL-7B: $l_0$=8; $l_{\delta}$=3
	\item InternVL2.5-8B: $l_0$=6; $l_{\delta}$=3
	\item InternVL2.5-26B: $l_0$=14; $l_{\delta}$=4
\end{itemize}

\subsection{More Visualisation of Low-Frequency Vision Tokens}

As illustrated in Fig.~\ref{fig:appendix_visual}, we test additional images and analyzed the visualization results of low-frequency visual tokens mapped back to the original images across different LLM layers of InternVL2.5-8B~\cite{internvl2_5}. Here, we employed the same prompt: "\textbf{Please describe the image shortly.}" The visualization results clearly demonstrate that low-frequency visual tokens encode richer visual semantics and retain more critical information, further supporting the reliability of the SPU strategy adopted in LUVC.

\end{document}

%% file: sec/0_abstract.tex
\begin{abstract}
	Visual language models encounter challenges in computational efficiency and latency, primarily due to the substantial redundancy in the token representations of high-resolution images and videos. Current attention/similarity-based compression algorithms suffer from either position bias or class imbalance, leading to significant accuracy degradation. They also fail to generalize to shallow LLM layers, which exhibit weaker cross-modal interactions. To address this, we extend token compression to the visual encoder through an effective iterative merging scheme that is orthogonal in spatial axes to accelerate the computation across the entire VLM. Furthermoer, we integrate a spectrum pruning unit into LLM through an attention/similarity-free low-pass filter, which gradually prunes redundant visual tokens and is fully compatible to modern FlashAttention. On this basis, we propose \textbf{L}ossless \textbf{U}ltimate \textbf{V}ision tokens \textbf{C}ompression~(\textbf{LUVC}) framework.  LUVC systematically compresses visual tokens until complete elimination at the final layer of LLM, so that the high-dimensional visual features are gradually fused into the multimodal queries. The experiments show that LUVC achieves a 2× speedup inference in language model with negligible accuracy degradation, and the training-free characteristic enables immediate deployment across multiple VLMs. 	
\end{abstract}